\documentclass[10pt,twocolumn,letterpaper]{article}

\usepackage[pagenumbers]{cvpr} 
\usepackage[dvipsnames]{xcolor}

\setlist[itemize]{itemsep=10pt, topsep=5pt}
\definecolor{cvprblue}{rgb}{0.21,0.49,0.74}
\newcommand{\modelname}{{Florence-VL}}

\usepackage[pagebackref,breaklinks,colorlinks,citecolor=cvprblue]{hyperref}
\usepackage{subcaption}
\usepackage{graphicx}
\usepackage{rotating}
\usepackage{colortbl} 
\usepackage{multirow}
\usepackage{ulem}
\def\paperID{5736} 
\def\confName{CVPR}
\def\confYear{2024}

\usepackage{etoolbox}
\AtBeginEnvironment{tabular}{\footnotesize} 

\title{\modelname: Enhancing Vision-Language Models with Generative Vision Encoder and Depth-Breadth Fusion}

\author{
Jiuhai Chen\textsuperscript{1}\thanks{The work is done during Jiuhai Chen’s internship at Microsoft Research.}, Jianwei Yang\textsuperscript{2}, Haiping Wu\textsuperscript{2}, Dianqi Li, Jianfeng Gao\textsuperscript{2}, Tianyi Zhou\textsuperscript{1}, Bin Xiao\textsuperscript{2} \\
    \textsuperscript{1}University of Maryland \\
    \textsuperscript{2}Microsoft Research \\
 \\
}

\begin{document}

\maketitle
\begin{abstract}

We present \modelname{}, a new family of multimodal large language models (MLLMs) with enriched visual representations produced by Florence-2~\cite{xiao2024florence}, a generative vision foundation model. Unlike the widely used CLIP-style vision transformer~\cite{radford2021learning} trained by contrastive learning, Florence-2 can capture different levels and aspects of visual features, which are more versatile to be adapted to diverse downstream tasks.  
We propose a novel feature-fusion architecture and an innovative training recipe that effectively integrates Florence-2's visual features into pretrained LLMs, such as Phi 3.5 and LLama 3. 
In particular, we propose ``depth-breath fusion (DBFusion)'' to fuse the visual features extracted from different depths and under multiple prompts. Our model training is composed of end-to-end pretraining of the whole model followed by finetuning of the projection layer and the LLM, on a carefully designed recipe of diverse open-source datasets 
that include high-quality image captions and instruction-tuning pairs. 
Our quantitative analysis and visualization of \modelname{}'s visual features show its advantages over popular vision encoders on vision-language alignment, where the enriched depth and breath play important roles. 
\modelname{} achieves significant improvements over existing state-of-the-art MLLMs across various multi-modal and vision-centric benchmarks covering general VQA, perception, hallucination, OCR, Chart, knowledge-intensive understanding, etc. To facilitate future research, our models and the complete training recipe are open-sourced. \href{https://github.com/JiuhaiChen/Florence-VL}{https://github.com/JiuhaiChen/Florence-VL}






\end{abstract}    
\section{Introduction}
\label{sec:intro}

\begin{figure}[ht]
\centering
\includegraphics[width=\linewidth]{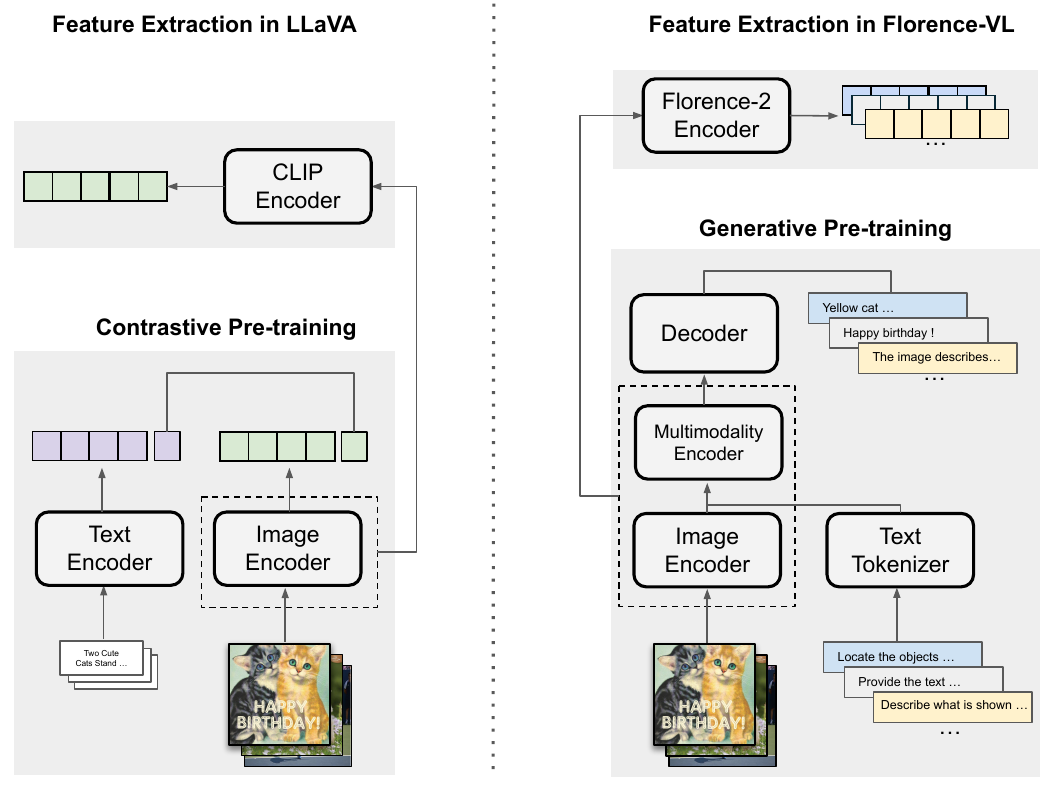}
\vspace{-1em}
\caption{Comparison of LLaVA-style MLLMs with our \modelname{}. LLaVA-style models use CLIP, pretrained with contrastive learning, to generate a \textbf{single high-level image feature}. In contrast, \modelname{} leverages Florence-2, pretrained with generative modeling across various vision tasks such as image captioning, OCR, and grounding. This enables \modelname{} to flexibly extract \textbf{multiple task-specific image features} using Florence-2 as the image encoder.}
\label{fig:teaser}
\vspace{-1em}
\end{figure}
Recent progress in multimodal large language models (MLLMs) are largely driven by progress in large language models \cite{liu2024visual, zhu2023minigpt}. However, when it comes to visual encoders, transformer-based models like CLIP or SigLIP remain the most commonly used choices. 
Despite CLIP and SigLIP’s effectiveness, they come with limitations; for instance, their last-layer features usually provide an image-level semantic representation that captures the overall scene and context, but often overlook pixel or region-level details and low-level features that are critical to various downstream tasks. 
There is a much broader range of visual representation, such as the self-supervised DINOv2 model \cite{oquab2023dinov2}, diffusion model \cite{rombach2022high} and segmentation \cite{kirillov2023segment}, \citep{tong2024cambrian} shows these different visual encoders can benefit well in some specific tasks. \looseness-1

\begin{figure*}[ht]
\centering
\includegraphics[width=0.8\textwidth]{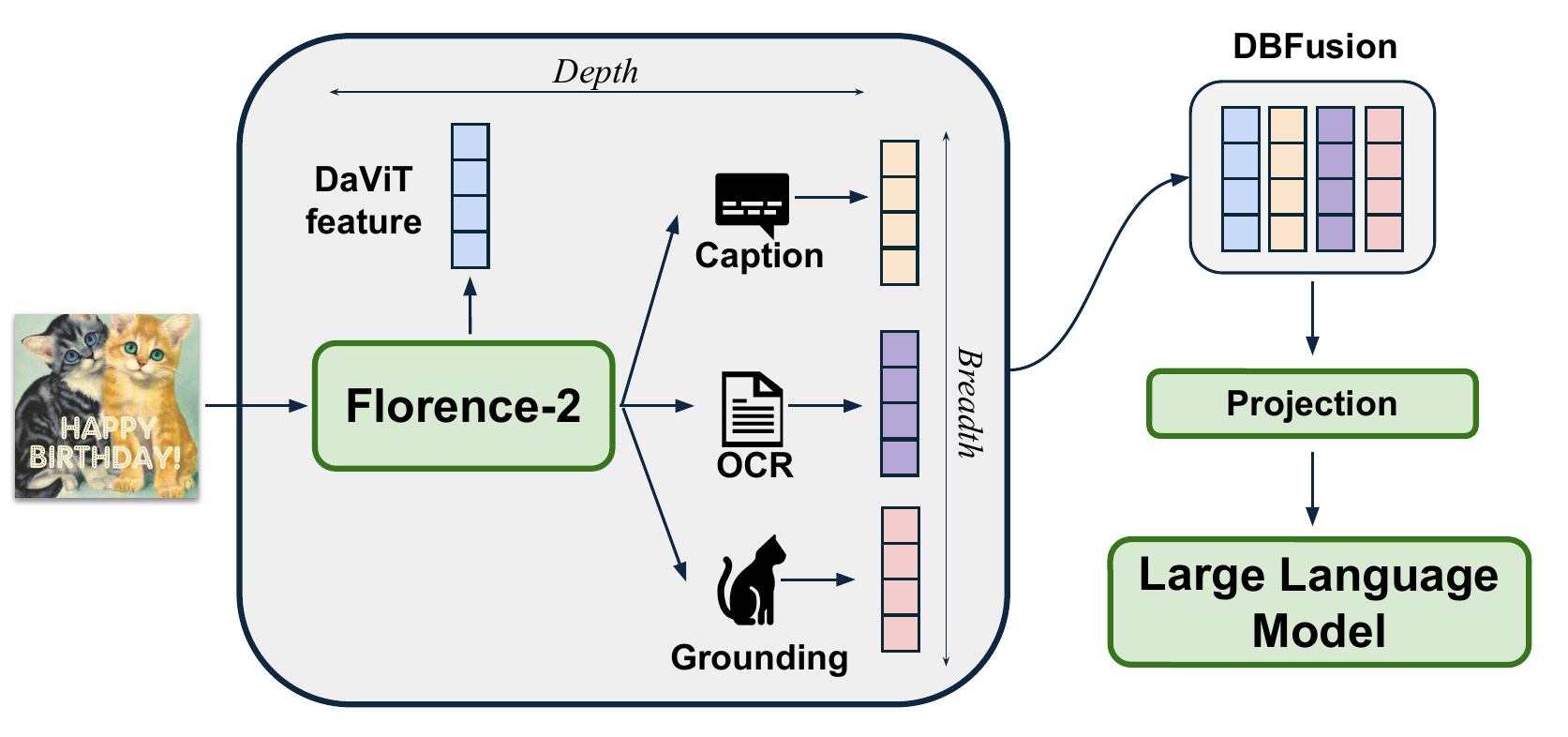}
\vspace{-1em}
\caption{An overview of \modelname{}, which extracts visual features of different depths (levels of feature concepts) and breaths (prompts) from Florence-2, combines them using DBFusion, and project the fused features to an LLM's input space. \modelname{} is fully pretrained on image captioning data and then partially finetuned on instruction-tuning data.}
\label{fig:method}
\end{figure*}


In order to leverage distinctive representations of multiple vision encoders, some recent works such as \cite{tong2024cambrian, shi2024eagle} adopt a mixture of vision encoders that specialize in different feature aspects or skills. However, integrating multiple vision encoders increases the computational expense for both model training and deployment. \textit{Could a single vision model be designed to generate distinct visual features, each emphasizing different perceptual information in the input image?} In this paper, we propose \modelname{}, which leverages the generative vision foundation model Florence-2 \cite{xiao2024florence} as the vision encoder. Florence-2 offers a prompt-based representation for various computer vision tasks, including captioning, object detection, grounding, and OCR. Its versatile visual representations can benefit different types of downstream tasks. For instance, OCR-based representations are advantageous for tasks that require extracting textual information from images, and grounding-based representation can benefit for tasks that require the relationships between objects and their spatial contexts. However, to build a better MLLM, how to extract these diverse features and align them with a pretrained LLM remains underexplored. 

To address this, we propose \textit{Depth-Breadth Fusion} (\textbf{DBFusion}) to effectively selecting and utilizing diverse visual features. Visual features from different layers capture various levels of concepts, with the final layers typically representing higher-level concepts. Integrating lower-level features can therefore complement these high-level representations, which we refer to as the ``Depth'' of visual features. Additionally, since different downstream tasks need different perceptual information within images, a single image feature often falls short in capturing all relevant information. Thus, we leverage multiple image features, with each feature capturing different visual representations. We refer to this as the “Breadth” of visual features. 
For utilizing these diverse visual features, we find that a straightforward channel concatenation serves as a simple yet effective fusion strategy. Specifically, we concatenate multiple features along the channel dimension, and these combined features, spanning various depths and breadths, are then projected as input embedding to LLMs.

We train \modelname{} on a novel recipe of open-sourced training data, which is composed of a large-scale detailed captioning dataset and a mix of instruction tuning datasets for whole-model pretraining and partial-model finetuning, respectively. 
The resulted \modelname{} achieves significant advantages on 25 benchmarks covering vision-centric, knowledge-based, and OCR \& Chart tasks, outperforming other advanced MLLMs like Cambrian \cite{tong2024cambrian}. Moreover, we provide quantitative analysis and visualization demonstrating that \modelname{}'s visual representation achieves better alignment to LLMs than the widely adopted vision encoders such as CLIP and SigLIP \cite{liu2024visual}.

\begin{figure*}[ht]
\centering
\includegraphics[width=0.9\textwidth]{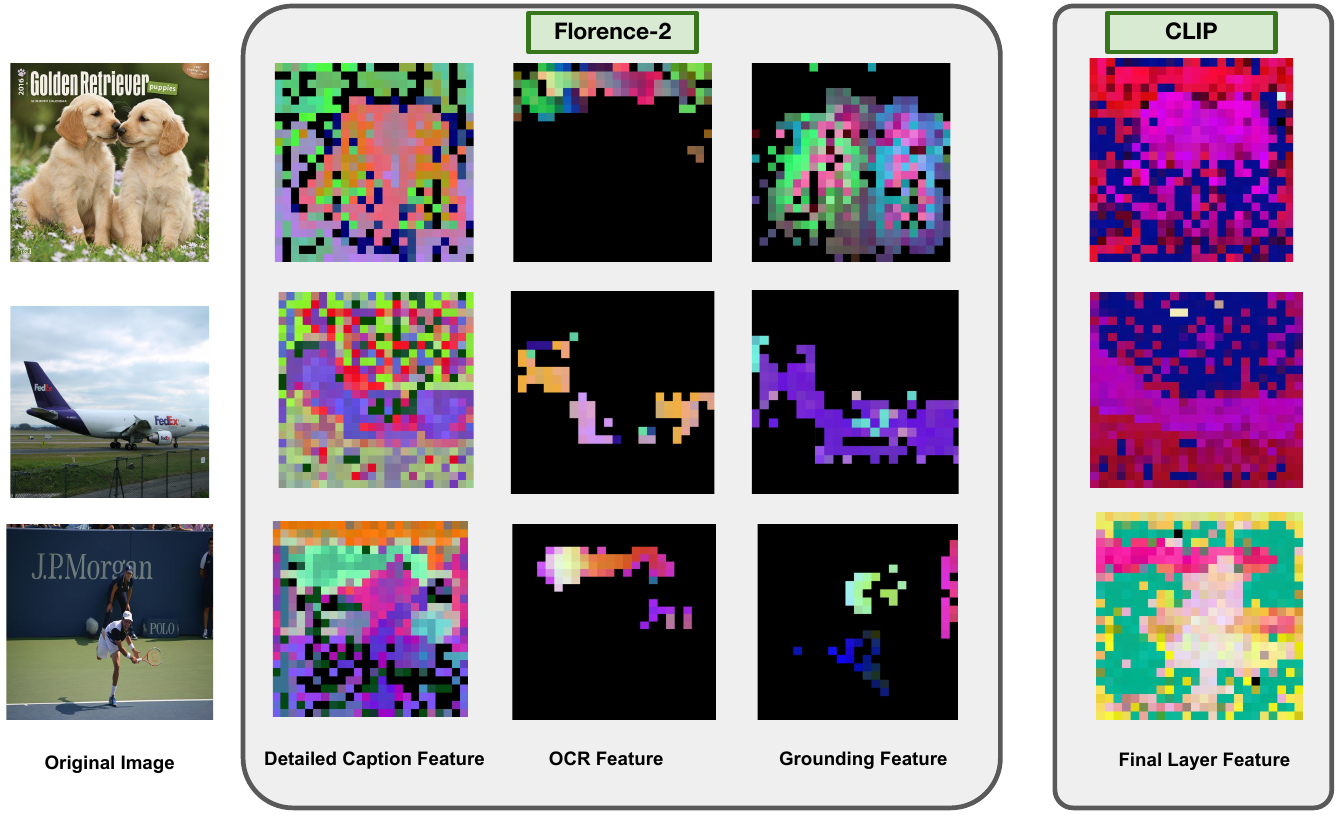}
\vspace{-1em}
\caption{Visualization of the first three PCA components: we apply PCA to image features generated from Detailed Caption, OCR, and Grounding prompts, excluding the background by setting a threshold on the first PCA component. The image features derived from the Detailed Caption prompt (second column) capture the general context of the image, those from the OCR prompt (third column) focus primarily on text information, and those from the Grounding prompt (fourth column) highlight spatial relationships between objects. Additionally, we visualize the final layer features from OpenAI CLIP (ViT-L/14@336) in the last column, showing that CLIP features often miss certain region-level details, such as text information in many cases. }
\label{fig:florence_feature}
\end{figure*}

\section{Preliminary: Florence-2}

Florence-2 \cite{xiao2024florence} is a vision foundation model that utilizes a unified, prompt-based approach to handle various vision tasks with simple instructions, such as captioning, object detection, grounding, and segmentation. The architecture consists of a vision encoder DaViT \cite{ding2022davit} and a standard encoder-decoder model. It processes an input image $\mathbf{I} \in \mathbb{R}^{H \times W \times 3}$ (where $H$ and $W$ indicate height and width, respectively) into flattened visual token embeddings.
The model then applies a standard encoder-decoder transformer architecture to process both visual and language token embeddings. It first generates prompt text embeddings $\mathbf{T} \in \mathbb{R}^{N_t \times D}$ using the language tokenizer and word embedding layer, with $N_t$ and $D$ representing the number and dimensionality of prompt tokens, respectively. The vision token embeddings  are then concatenated with the prompt embeddings to create the input for the multi-modality encoder module, $\mathbf{X} = [\mathbf{V}, \mathbf{T}]$, where $\mathbf{V} \in \mathbb{R}^{N_v \times D}$ is produced by applying a linear projection and LayerNorm layer to visual embedding from DaViT, with $N_v$ and $D$ representing the number and dimensionality of vision tokens, respectively. The linear projection and LayerNorm layer are used to ensure dimensionality alignment with $\mathbf{T}$. Encoder-decoder model will process the $\mathbf{X}$ and generate the desirable results, such as captions, object detections, grounding in textual form.


\section{Method}
\label{sec:method}

\subsection{Using Florence-2 as Vision Backbone} To address the limitations of existing vision backbones in MLLMs, specifically, last layer features typically yield an image-level representation that captures overall scene and context but often misses pixel- or region-level details, we utilize the vision foundation model Florence-2 as our visual encoder for extracting visual features. Unlike the CLIP pre-trained vision transformers that provide a single, universal image feature, Florence-2 can identify spatial details at different scales, by using different tasks prompts. 

In MLLMs, effective image understanding requires capturing multiple levels of granularity, from global semantics to local details, and understanding spatial relationships between objects and entities within their semantic context.
Florence-2, with its capability to manage diverse granularity levels, is an ideal vision encoder to address these core aspects of image comprehension. In the following section, we explore how to leverage Florence-2's strengths in integrating it into MLLMs.

\begin{table*}[ht]
    \centering
    \resizebox{1.0\textwidth}{!}{
        \begin{tabular}{cc|cccccccccccccc|c}
           & \rotatebox{90}{\# Vis tok} & \rotatebox{90}{ MMBench (EN)} & \rotatebox{90}{POPE} & \rotatebox{90}{MM-Vet} & \rotatebox{90}{MME-P} & \rotatebox{90}{Seed-image} & \rotatebox{90}{HallusionBench} & \rotatebox{90}{LLaVA-bench}  & \rotatebox{90}{AI2D} & \rotatebox{90}{MathVista} & \rotatebox{90}{MMMU} & \rotatebox{90}{OCRBench} & \rotatebox{90}{ChartQA} & \rotatebox{90}{DocVQA} & \rotatebox{90}{InfoVQA} & \rotatebox{90}{Average}\\
         \hline
         \hline  
    Token Integration & 1728  & \textbf{66.6}  & 88.7 & 34.1 & 1536.3 & \textbf{70.9} & 45.0 & 63.3 & 56.9 & \textbf{28.1} & \textbf{36.4}  & 40.8 & 23.0 & 44.6 & \textbf{29.5} & 50.3 \\
    Average Pooling & 576  & 65.7 & 88.8 & 32.3 & \textbf{1551.3}  & 70.3 & 45.7 & 64.6 & 56.6 & 27.4 & 36.0  & 41.2 & \textbf{24.6} & \textbf{44.8} & 29.3 & 50.4 \\
    Channel Integration &576  & 66.1 & \textbf{89.4} & \textbf{35.2} & 1543.5 & 70.3 & \textbf{46.8} & \textbf{65.0} & \textbf{57.2} & 28.0 & 35.6 & \textbf{41.4} & 24.3 & 44.5 & 29.4 & \textbf{50.8} \\
        \hline
        \hline
        \end{tabular}
    }
    \caption{Experiments for different fusion strategies. The vision token count is 1728 for token integration, which leads to longer training and inference times. The channel integration strategy shows better performance and training efficiency compared to the other two fusion methods.}
    \label{tab:feature_fusion_ablation}
\end{table*}

\subsection{Visual Features spanning Depth and Breadth}

\paragraph{Breadth.} Since different downstream tasks require varying perceptual information from images, we consider expanding the breadth of visual representation. Given an input image $\mathbf{I} \in \mathbb{R}^{H \times W \times 3}$ and a task-specific prompt, such as "provide the text shown in the image", Florence-2 will process the image feature and prompt feature into $\mathbf{X} = [\mathbf{V}, \mathbf{T}]$ and then feed into the encoder-decoder transformer architecture. The encoder employs an attention mechanism to process $\mathbf{X}$, producing an output $\mathbf{X}' = [\mathbf{V}', \mathbf{T}']$. Due to the cross-attention between $\mathbf{V}$ and $\mathbf{T}$, the updated image feature $\mathbf{V}'$ becomes more focused on the prompt "provide the text shown in the image", specifically extracting more text information from the image.

We focus on three distinct tasks that contribute to image understanding, resulting in three different image embeddings $[\mathbf{V}_{t_1}', \mathbf{V}_{t_2}', \mathbf{V}_{t_3}']$, each tailored to a specific task:

\begin{itemize}[itemsep=1pt]
  \item \textbf{Detailed Image Caption}: describe what is shown in the image  with a paragraph. It enables the model to give a overall context of an image.
    \item \textbf{OCR}: provide the text shown in the image. It extracts more text information from the image.
  \item \textbf{Dense Region Caption}: locate the objects in the image, with their descriptions. It captures the spatial relationships between objects.
\end{itemize}

We visualize the image features with different task prompts, applying PCA to the visual embeddings and setting a threshold for the visualization. As illustrated in Figure \ref{fig:florence_feature}, different image embeddings emphasize distinct conceptual information within the images. Additionally, we also visualize the final layer image features from OpenAI CLIP in Figure \ref{fig:florence_feature}, which often lacks certain region-level details in most cases.


\paragraph{Depth.} 
We also integrate lower-level features using $\mathbf{V}$ from DaViT, combined with higher-level features $[\mathbf{V}_{t_1}', \mathbf{V}_{t_2}', \mathbf{V}_{t_3}']$ derived from the three prompts, allows us to capture multiple levels of conceptual detail.



\subsection{Depth-Breadth Fusion}

Since we have image feature with different level of granularity, feature fusion is commonly used. When dealing with multiple feature embeddings, such as $[\mathbf{V}, \mathbf{V}_{t_1}', \mathbf{V}_{t_2}', \mathbf{V}_{t_3}']$, the next question becomes how to fuse these features and align them with the language model space. To take advantage of all these four features, several approaches can be considered for this fusion process:

\begin{itemize}[itemsep=1pt]
  \item \textbf{Token Integration}: This approach involves concatenating all features along the token dimension. However, this can make the visual token excessively long and complicate model training.
\item \textbf{Average Pooling}: Alternatively, average pooling over all features can be used, but this method may result in information loss.
  \item \textbf{Channel Integration}: A more effective method is to concatenate features along the channel dimension, which does not increase the sequence length.
\end{itemize}


To quickly assess which feature fusion method provides the best overall performance, we use datasets from LLaVA-1.5 \cite{liu2024visual}, which include 558K image captions for pre-training and 665K entries for instruction tuning. In the Table \ref{tab:feature_fusion_ablation}, the channel integration strategy shows better performance and training efficiency compared to the other two fusion methods. Thus we choose channel integration simple yet effective fusion strategy.

\subsection{\modelname{}}
As shown in Figure \ref{fig:method}, \modelname{} is composed of the vision foundation model Florence-2 and the large language model. After extracting multiple image features, we use MLP to project these features into the language model space. During the pretraining stage, we align Florence-2 with the language model using image detailed caption data. In the instruction tuning stage, we use diverse and high-quality instruction-tuning dataset to effectively adapt the model to downstream tasks.

\section{Analysis on Different Vision Encoders}

To demonstrate that Florence-2 is a superior vision encoder compared to others, we quantify the cross-modal alignment quality between various vision encoders and language models, allowing us to assess the impact of different vision encoders without requiring subsequent supervised fine-tuning and evaluations on benchmarks \cite{huang2024deciphering, wei2024large}. Specifically, consider a pretrained MLLM $\mathcal{M} = (\mathcal{V}, \mathcal{L})$ where $\mathcal{V}$ is the vision encoder and $\mathcal{L}$ represents the language model, we input a set of image-text pairs, $(V, T) = (\{v_n\}_{n=1}^N, \{t_n\}_{n=1}^N)$, into the model. 
For the $n^{th}$ image-text pair, the vision encoder produces vision representations $f^{v_n} \in \mathbb{R}^{r_n \times d'}$, and the text representations $f^{t_n} \in \mathbb{R}^{s_n \times d}$ from last layer of the language decoder, 
where $r_n$ and $s_n$ are the number of tokens in the vision and text representations, and $d'$ and $d$ are the hidden state dimensions for the vision and text tokens. We apply the trainable projection $\mathcal{P}$ to $f^{v_n}$ to ensure dimensionality alignment with $f^{t_n}$, that is $\mathcal{P}(f^{v_n}) \in \mathbb{R}^{r_n \times d}$. We also apply average pooling along token dimension and normalize along the hidden dimension for both $\mathcal{P}(f^{v_n})$ and $f^{t_n}$. 
For all image-text pairs, we concatenate all vision features along the first dimension to form a matrix $F^{v_n} \in \mathbb{R}^{N \times d}$, and similarly concatenate all text features into a matrix $F^{t_n} \in \mathbb{R}^{N \times d}$. 
Since we need to measure the modality gap between vision tokens and text tokens, we compute the divergence between these two token representations. Specifically, we optimize the trainable projection $\mathcal{P}$, which is used to bring these two representations closer together by minimizing a cross-entropy loss function: 
\[
\mathcal{L} = -\sum_{i,j} \mathcal{I}_n^{(i,j)} \log \left( \text{softmax}(F^{v_n} \times (F^{t_n})^T)_{i,j} \right)
\], where $\mathcal{I}_n$ is the target (indicator) matrix. The multiplication of $F^{v_n}$ with the transpose of $F^{t_n}$ calculates the correlation between vision and text token representations.
In short, the loss function is designed to minimize the distance between vision tokens and their corresponding text tokens by maximizing the likelihood that each vision token aligns correctly with its associated text token.

\begin{figure}[ht]
\centering
\includegraphics[width=\linewidth]{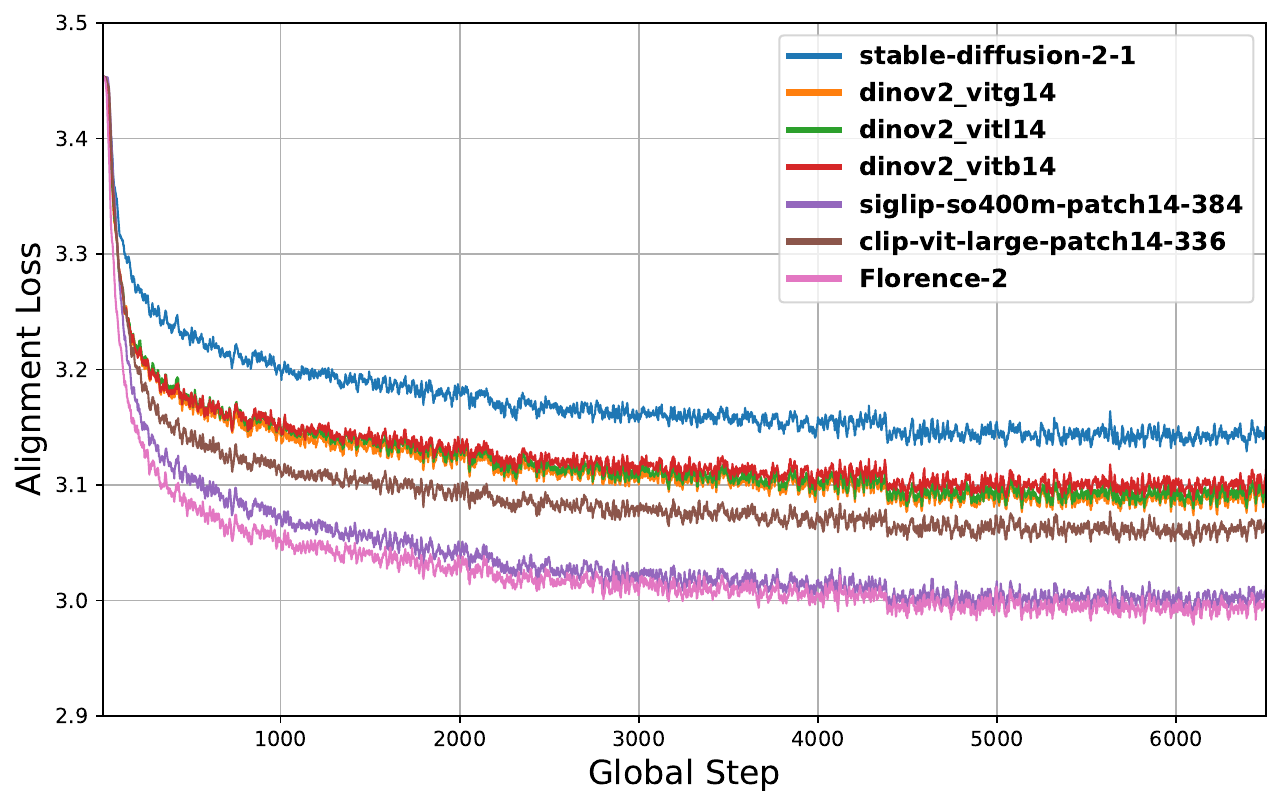}
\vspace{-1em}
\caption{We plot the alignment loss for different vision encoders, which clearly shows that Florence-2 vision encoder achieves the lowest alignment loss compared to the other vision encoders, demonstrating the best alignment with text embeddings. \label{fig:alignment_loss}}
\end{figure}

We use a set of image-text pairs $(V, T) = (\{v_n\}_{n=1}^N, \{t_n\}_{n=1}^N)$ from the LLaVA 1.5 pretraining image captioning datasets and select various vision encoders to assess how well we can optimize the alignment between the vision encoder and the language model. The vision encoders we evaluate include: Stable Diffusion \cite{Rombach_2022_CVPR}, Dinov2 \cite{oquab2023dinov2} (ViT-G/14, ViT-L/14, ViT-B/14), SigLIP, OpenAI CLIP, and our Florence-2 model. The chosen language model is Llama 3 8B Instruct. We plot the alignment loss in Figure \ref{fig:alignment_loss}, which clearly shows that Florence-2 vision encoder achieves the lowest alignment loss compared to the other vision encoders, demonstrating the best alignment with text embeddings. Additionally, SigLIP demonstrates competitive results, as noted in \cite{tong2024cambrian}, which highlights SigLIP's strong benchmark performance relative to other vision encoders, aligning with the findings of our study.

\begin{figure}[ht]
\centering
\includegraphics[width=\linewidth]{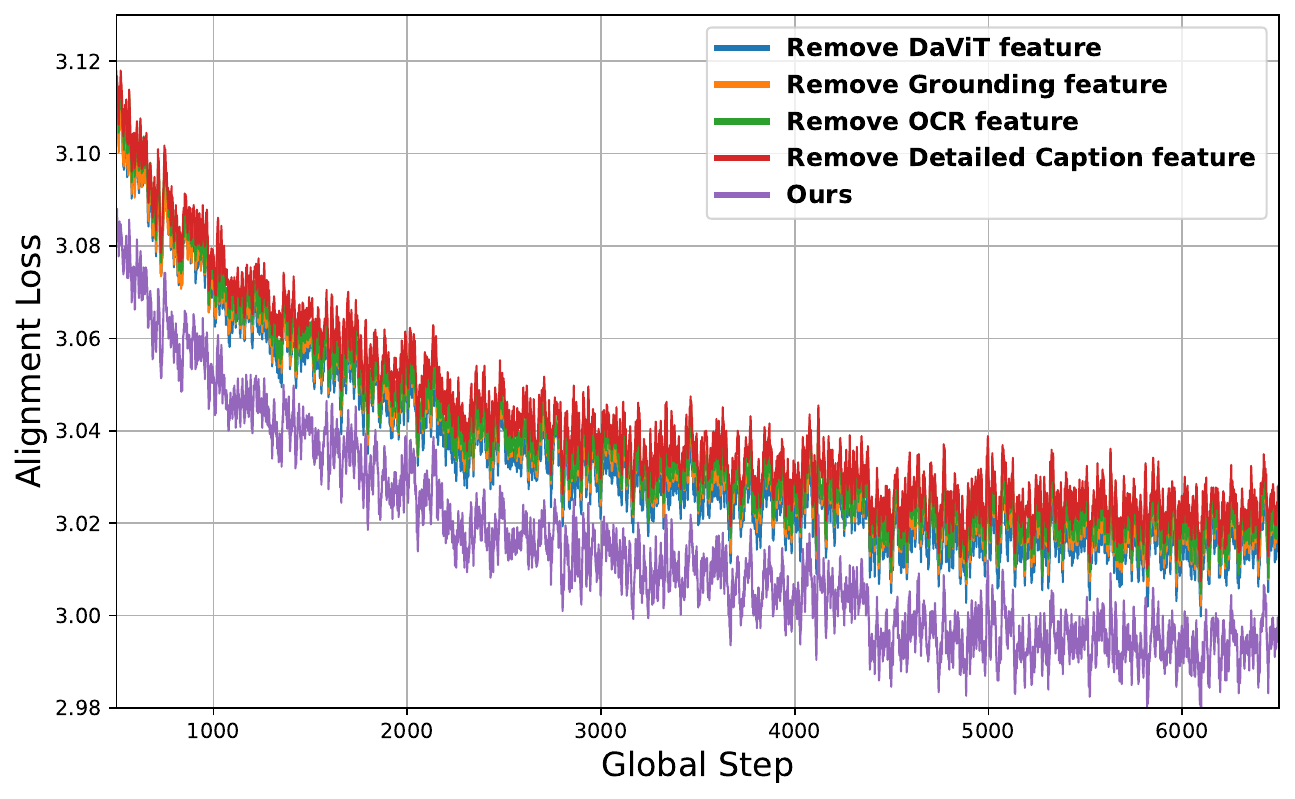}
\vspace{-1em}
\caption{We plot the alignment loss for various feature combinations, removing one feature at a time from different depths and breadths. The results clearly show that our method achieves the lowest alignment loss compared to others, highlighting the importance of all features from different depths and breadths for optimal alignment.\label{fig:feature_remove_loss}}
\end{figure}

\begin{table*}[ht]
    \centering
    \begin{subtable}[t]{\textwidth}
        \centering
        \begin{tabular}{cc|cccccccccccccc}
        & & \multicolumn{13}{c}{General Benchmarks}\\
         & \rotatebox{90}{\# Vis tok.} & \rotatebox{90}{VQAv2} & \rotatebox{90}{GQA} & \rotatebox{90}{MMBench (EN)} & \rotatebox{90}{MMBench (CN)} & \rotatebox{90}{VizWiz} & \rotatebox{90}{POPE} & \rotatebox{90}{MM-Vet} & \rotatebox{90}{MME-P} & \rotatebox{90}{MME-C}  & \rotatebox{90}{Seed-image} & \rotatebox{90}{HallusionBench} & \rotatebox{90}{LLaVA-bench} & \rotatebox{90}{MMStar} \\
         \hline
         \hline
        Vila 3B & - & 80.4 & 61.5 & 63.4 & 52.7 & 53.5 & 86.9 & 35.4 & 1442.4 & - & 67.9 & - & - & 40.3 \\
        Phi 3.5-Vision & - & - & \textbf{63.5} & \textbf{75.5} & \textbf{64.2} & 58.2 & 82.2 & 46.5 & 1473.4 & \textbf{412.1} & 69.9 & 53.3 & 68.8 & \textbf{49.0} \\
        \rowcolor{lightgray} \modelname{} 3B (ours) & 576 & \textbf{82.1} & 61.8 & 71.6 & 60.8 & \textbf{59.1} & \textbf{88.3} & \textbf{51.0} & \textbf{1498.7} & 403.9 & \textbf{70.6} & \textbf{58.1} & \textbf{71.1} & 44.9 \\
        \hline
        \hline
        LLaVA next 8B & 2880 & - & \textbf{65.4} & 72.2 & - & 57.7 & 86.6 & 41.7 & 1595.1 & 379.3 & 72.7 & 47.7 & \textbf{76.8} & - \\
        Vila 8B & - & 80.9 & 61.7 & 72.3 & 66.2 & 58.7 & 84.4 & 38.3 & 1577.0 & - & 71.4 & - & - & - \\
        Mini-Gemini-HD 8B & 2880 & - & 64.5 & 72.7 & - &  - & - & - & \textbf{1606.0} & - & 73.2 & - & - & -  \\
        Cambrain 8B & 576 & - & 64.6 & 75.9 & 67.9 & - & 87.4 & 48.0 & 1547.1 & - &  74.7 & 48.7 & 71.0 & \textbf{50.0} \\
        \rowcolor{lightgray} \modelname{} 8B (ours) & 576 & \textbf{84.7} & 64.4 & \textbf{76.2} & \textbf{69.5} & \textbf{59.1} & \textbf{89.9} & \textbf{56.3} & 1560.0 & \textbf{381.1} & \textbf{74.9} & \textbf{57.3} & 74.2 & \textbf{50.0} \\
        \hline
        \hline
        \end{tabular}%
        \caption{Results on general multimodal benchmarks.}
        \label{tab:general_benchmark}
    \end{subtable}%
    \vspace{1em} 
    \begin{subtable}[t]{\textwidth}
        \centering
        \begin{tabular}{cc|ccc|cccc|cccccccc}
        & & \multicolumn{3}{c}{Vision centric} & \multicolumn{4}{c}{Knowledge based} & \multicolumn{5}{c}{OCR \& Chart}\\
         & \# Vis tok. & \rotatebox{90}{Realworldqa} & \rotatebox{90}{CV-Bench*} & \rotatebox{90}{MMVP} & \rotatebox{90}{AI2D} & \rotatebox{90}{MathVista} & \rotatebox{90}{MMMU} & \rotatebox{90}{SciQA-IMG} & \rotatebox{90}{TextVQA} & \rotatebox{90}{OCRBench} & \rotatebox{90}{ChartQA} & \rotatebox{90}{DocVQA} & \rotatebox{90}{InfoVQA} \\
         \hline
         \hline
        Vila 3B & - & 53.3  & 55.2 & - & - & 30.6 & 34.1 & 67.9 & 58.1 & - & - & - & - \\
        Phi 3.5 Vision & - &  53.5  & 69.3 & \textbf{67.7} & \textbf{77.4} & - & \textbf{43.3} & \textbf{89.0} & 61.1 & 59.8 & \textbf{72.0} & 75.9  & 40.7 \\
        \rowcolor{lightgray} \modelname{} 3B (ours) & 576 &  \textbf{60.4}  & \textbf{70.2}  & 64.7 & 73.8 & \textbf{52.2} & 41.8 & 84.6 & \textbf{69.1} & \textbf{63.0} & 70.7 & \textbf{82.1} & \textbf{51.3} \\
         \hline
         \hline
        LLaVA next 8B & 2880 &  59.6 & 63.8 & 38.7 & 71.6 & 37.4 & 40.1 & 73.3 & 65.4 & 55.2 & 69.3 & 78.2 & - \\
        Vila 8B & - &  - & - & - & - & - & 36.9 & 79.9 & - & - & - & - & - \\
        Mini-Gemini-HD 8B & 2880 & 62.1 & 62.6 & 18.7 & 73.5 & 37.0 & 37.3 & 75.1 & 70.2 & 47.7 & 59.1 & 74.6 & -  \\
        Cambrian 8B & 576 & \textbf{64.2} & 72.2 & 51.3 & 73.0 & 49.0 & 42.7 & 80.4 & 71.7 & 62.4 & 73.3 & 77.8 & - \\
        \rowcolor{lightgray} \modelname{} 8B (ours) & 576 & \textbf{64.2} & \textbf{73.4} & \textbf{73.3} & \textbf{74.2} & \textbf{55.5} & \textbf{43.7} & \textbf{85.9} & \textbf{74.2} & \textbf{63.4} & \textbf{74.7} & \textbf{84.9} & \textbf{51.7} \\
         \hline
         \hline
        \end{tabular}%
        \caption{Results on Vision centric, Knowledge based, and OCR \& Chart benchmarks.}
    \end{subtable}
    \caption{\label{tab:combined_benchmarks}
Results on general multimodal benchmarks, Vision centric, Knowledge based, and OCR \& Chart benchmarks.}
\end{table*}

\section{Experiments}
\paragraph{Implementation Details.} 


In order to build a state-of-the-art MLLM, we use images from CC12M \cite{changpinyo2021conceptual}, Redcaps \cite{desai2021redcaps}, and Commonpool \cite{gadre2024datacomp} during the pretraining stage, with detailed captions sourced from PixelProse \cite{singla2024pixelsproselargedataset}. For the instruction tuning stage, we also curate our high quality instruction tuning datasets, sourcing from Cambrian-7M \cite{tong2024cambrian}, Vision Flan \cite{xu2024visionflanscalinghumanlabeledtasks}, ShareGPT4V \cite{chen2023sharegpt4vimprovinglargemultimodal}, along with additional data from Docmatix \cite{Docmatix} to improve chart and diagram comprehension \cite{bai2023qwenvlversatilevisionlanguagemodel}. The detail of training datasets and experiment details can be found in the appendix.

\paragraph{Evaluation.} We evaluate the performance of different MLLM models on 25 benchmarks with four different categories:
\begin{itemize}[itemsep=0.5pt]
\item  General multimodal benchmarks: VQAv2~\cite{goyal2017making}, GQA~\cite{hudson2019gqa}, MMBench (EN and CN)~\cite{liu2023mmbench}, VisWiz~\citep{gurari2018vizwiz}, POPE~\cite{li2023evaluating}, MM-Vet~\citep{yu2023mm}, MME Perception~\cite{fu2024mme}, MME Cognition~\cite{fu2024mme}, SeedBench~\citep{li2023seed}, HallusionBench,  LLaVA in the Wild~\cite{liu2024visual} and MMStar~\cite{chen2024we}.

\item  OCR \& Chart benchmark: TextVQA~\cite{singh2019towards}, OCRBench~\cite{liu2024hidden}, ChartQA~\cite{masry2022chartqa}, DocVQA~\cite{mathew2021docvqa} and InforVQA~\cite{mathew2022infographicvqa}.

\item  Knowledge based benchmark: AI2D~\cite{kembhavi2016diagram}, MathVista~\cite{lu2023mathvista}, MMMU~\cite{yue2024mmmu} and ScienceQA~\cite{lu2022learn}.

\item  Vision Centric benchmark: MMVP~\cite{tong2024eyes}, RealworldQA~\cite{grok15v} and CV-Bench~\cite{tong2024cambrian}.
\end{itemize}

\paragraph{Baselines.}  We select two language backbones: Phi-3.5-mini-Instruct and LLama-3-8B-Instruct. For baseline comparisons among small models, we chose Vila 1.5 3B \cite{lin2024vila} and Phi 3.5-Vision-Instruct \cite{abdin2024phi}. For the larger models, we select the baselines: LLaVA Next 8B \cite{liu2024llavanext}, Vila 8B \cite{lin2024vila}, Mini-Gemini-HD 8B \cite{li2024mini} and Cambrain 8B \cite{tong2024cambrian}, using LLama 3 8B Instruct as the language backbone.

\begin{table*}[ht]
    \centering
    \begin{subtable}[t]{\textwidth}
        \centering
        \begin{tabular}{cc|cccccccccccc}
         & LLM  & \rotatebox{90}{GQA} & \rotatebox{90}{MMBench (EN)} & \rotatebox{90}{MMBench (CN)} & \rotatebox{90}{VizWiz} & \rotatebox{90}{POPE} & \rotatebox{90}{MM-Vet} & \rotatebox{90}{MME-P} & \rotatebox{90}{MME-C}   & \rotatebox{90}{HallusionBench} & \rotatebox{90}{LLaVA-bench} & \rotatebox{90}{MMStar} \\
         \hline
         \hline
            LLaVA 1.5 3B &  Phi 3.5 & 61.4 & \textbf{69.4} & 60.6 & 38.4 & 86.2 & 35.4 & 1399.5 & 284.6  & 44.5 & \textbf{68.0} & 40.6 \\         \modelname{} 3B & Phi 3.5 & \textbf{62.7} & 68.7 & \textbf{61.7} & \textbf{42.6} & \textbf{89.9} & 35.4 & \textbf{1448.5} & \textbf{299.6}  & \textbf{45.5} & 64.9 & \textbf{40.8} \\	
        \hline
        \hline
        LLaVA 1.5 7B & Vicuna 1.5  & 62.0 & 64.8 & \textbf{57.6} & 50.0 & 85.9 & 30.6 & 1510.7 & 294.0  & 44.8 & 64.2 & 30.3 \\	
        \modelname{} 7B & Vicuna 1.5  & \textbf{62.7} & \textbf{66.1} & 55.8 & \textbf{54.5} & \textbf{89.4} & \textbf{35.2} & \textbf{1543.5} & \textbf{316.4}  & \textbf{46.8} & \textbf{65.0} & \textbf{36.8} \\
        \hline
        \hline
        LLaVA 1.5 8B & Llama 3  & 62.8 & \textbf{71.4} & 65.5 & 49.3 & 84.8 & 34.2 & 1539.4 & 292.5 & 45.7 & \textbf{71.0} & 38.5 \\
         \modelname{} 8B & Llama 3  & \textbf{63.8} & 71.1 & \textbf{65.8} & \textbf{54.0} & \textbf{88.4} & \textbf{36.4} & \textbf{1584.1} & \textbf{346.8} & \textbf{46.8} & 66.2 & \textbf{39.1} \\
        \hline
        \hline
        \end{tabular}
        \caption{Results on general multimodal benchmarks. }
        \label{tab:ablation_method_11}
    \end{subtable}%
    \vspace{1em} 
    \begin{subtable}[t]{\textwidth}
        \centering
        \begin{tabular}{cc|cc|cccc|cccccccc}
         & LLM & \rotatebox{90}{Realworldqa}  & \rotatebox{90}{MMVP} & \rotatebox{90}{AI2D} & \rotatebox{90}{MathVista} & \rotatebox{90}{MMMU} & \rotatebox{90}{SciQA-IMG} & \rotatebox{90}{TextVQA} & \rotatebox{90}{OCRBench} & \rotatebox{90}{ChartQA} & \rotatebox{90}{DocVQA} & \rotatebox{90}{InfoVQA} \\
         \hline
         \hline
         LLaVA 1.5 3B & Phi 3.5 & 54.4 & 2.0 & 63.3 & 30.6 &  \textbf{40.7} & \textbf{72.0} & 43.7 & 30.4 & 16.4 & 28.1 & 26.4 \\	
        \modelname{} 3B & Phi 3.5  & \textbf{58.4}  & \textbf{6.0} & \textbf{64.9} & 30.6 & 39.6 & 68.7 & \textbf{61.6} & \textbf{40.3} & \textbf{21.8} & \textbf{46.1} & \textbf{29.6} \\	
        \hline
        LLaVA 1.5 7B & Vicuna 1.5 & 54.8   & 6.0 & 54.8 & 26.7 & 35.3 & \textbf{66.8} & 58.2 & 31.4 & 18.2 & 28.1 & 25.8 \\	
        \modelname{} 7B & Vicuna 1.5  & \textbf{60.4}   & \textbf{12.3} & \textbf{57.2} & \textbf{28.0} & \textbf{35.6} & 66.5 & \textbf{62.8} & \textbf{41.4} & \textbf{24.3} & \textbf{44.5} & \textbf{29.4} \\
         \hline
        LLaVA 1.5 8B & Llama 3 & 55.7  & 7.3 & 60.2 & 29.3 & 39.4 & \textbf{76.5} & 45.4 & 34.6 & 15.4 & 28.6 & 26.4 \\
        \modelname{} 8B & Llama 3 &  \textbf{59.9}  &  \textbf{8.3} & \textbf{62.4} & \textbf{31.8} & \textbf{39.9} & 73.6 & \textbf{68.0} & \textbf{41.1} & \textbf{23.4} & \textbf{44.4} & \textbf{29.0} \\
         \hline
         \hline
        \end{tabular}%
        \caption{Results on Vision centric, Knowledge based, and OCR \& Chart benchmarks.}
        \label{tab:ablation_method_12}
    \end{subtable}
    \caption{\label{tab:ablation_method_1}We compare LLaVA 1.5 with our model (\modelname{} 3B/7B/8B) across multiple multimodal benchmarks. The key difference between them lies in the vision encoders used (CLIP for LLaVA vs. Florence-2 for our model), while we maintain the same training data and backbone LLMs for both. The results show that our models significantly outperform LLaVA 1.5 with the same training data.}
\end{table*}

\paragraph{Results.} In the Table \ref{tab:combined_benchmarks}, we present the results of \modelname{} compared to various baselines across a range of benchmarks, along with the number of visual tokens used. For the smaller-sized model, our model outperforms Vila 3B, and surpasses Phi 3.5 Vision on 12 out of 24 tasks. Notably, Phi 3.5 Vision utilizes 500 billion vision and text tokens \cite{abdin2024phi}, with its training data being proprietary and significantly larger than ours. Nonetheless, our \modelname{} 3B remains competitive with this model.
For the larger-sized model, our model shows a significant improvement over other baselines on most benchmarks. Notably, our model significantly outperforms Cambrain-8B, which utilizes multiple vision encoders and combines their image features, whereas we achieve superior results using only a single vision encoder. 


\section{Discussion}

\paragraph{Results using LLaVA 1.5 Data.} Since we curate our training data when building our MLLMs, we disentangle the effects of training data and model architecture to clearly demonstrate our method's effectiveness. Specifically, to highlight the advantages of our model architecture, we use the \textbf{exact same} pretraining and instruction dataset as LLaVA 1.5 \cite{liu2024visual}. We test different language backbones, including Phi-3.5-mini-Instruct, Vicuna 1.5 7B, and LLama-3-8B-Instruct. As shown in Tables \ref{tab:ablation_method_1}, our model design significantly outperforms the LLaVA architectures when trained on the same dataset. Notably, for OCR \& Chart tasks, \modelname{} significantly outperforms LLaVA 1.5, demonstrating that OCR image features are essential for effective text-based image understanding.

\paragraph{Study on Depth Features Impacts.}

We aim to examine the impact of image features from different depths. For the feature set $[\mathbf{V}, \mathbf{V}_{t_1}', \mathbf{V}_{t_2}', \mathbf{V}_{t_3}']$, we first remove all higher-level features $[\mathbf{V}_{t_1}', \mathbf{V}_{t_2}', \mathbf{V}_{t_3}']$ and retain only the lower-level feature $[\mathbf{V}]$. We then evaluate the performance across different benchmarks, and as shown in Table \ref{tab:depth_feature_ablation}, using only the lower-level feature $[\mathbf{V}]$ performs worse than our complete method. Next, we remove the lower-level feature $[\mathbf{V}]$ and keep only the higher-level features $[\mathbf{V}_{t_1}', \mathbf{V}_{t_2}', \mathbf{V}_{t_3}']$. The alignment loss, displayed in Figure \ref{fig:feature_remove_loss}, clearly indicates that excluding the lower-level features (i.e., removing DaViT features) results in a higher alignment loss compared to our method. Therefore, both ablation studies confirm that features from different depths are essential for achieving optimal performance.

\begin{table*}[ht]
    \centering
        \begin{tabular}{c|cccccccccccccc}
           Features used & \rotatebox{90}{ MMBench (EN)} & \rotatebox{90}{POPE} & \rotatebox{90}{MM-Vet} & \rotatebox{90}{MME-P} & \rotatebox{90}{Seed-image} & \rotatebox{90}{HallusionBench} & \rotatebox{90}{LLaVA-bench}  & \rotatebox{90}{AI2D} & \rotatebox{90}{MathVista} & \rotatebox{90}{MMMU} & \rotatebox{90}{OCRBench} & \rotatebox{90}{ChartQA} & \rotatebox{90}{DocVQA} & \rotatebox{90}{InfoVQA} \\
         \hline
         \hline  
    $[\mathbf{V}]$  & 64.3 & 86.1 & 31.1 & 1510.7 & 66.0 & 44.8 & 64.2 & 54.7 & 26.7 & 35.2  & 31.2 & 18.3 & 27.9 & 25.7  \\
    $[\mathbf{V}, \mathbf{V}_{t_1}', \mathbf{V}_{t_2}', \mathbf{V}_{t_3}']$  & \textbf{66.1} & \textbf{89.4} & \textbf{35.2} & \textbf{1543.5} & \textbf{70.3} & \textbf{46.8} & \textbf{65.0} & \textbf{57.2} & \textbf{28.0} & \textbf{35.6} & \textbf{41.4} & \textbf{24.3} & \textbf{44.5} & \textbf{29.4} \\
        \hline
        \hline
        \end{tabular}
    \caption{The comparison between keeping only the lower-level feature $[\mathbf{V}]$ and our method, which includes both lower- and higher-level features, clearly demonstrates that maintaining both types of features achieves better performance.}
    \label{tab:depth_feature_ablation}
\end{table*}

\begin{table*}[ht]
    \centering
        \begin{tabular}{c|cccccccccccc|c}
           & \rotatebox{90}{GQA} & \rotatebox{90}{MMBench (EN)} & \rotatebox{90}{MMBench (CN)} & \rotatebox{90}{VizWiz} & \rotatebox{90}{POPE} & \rotatebox{90}{MM-Vet} & \rotatebox{90}{MME-P} & \rotatebox{90}{MME-C}  & \rotatebox{90}{Seed-image} & \rotatebox{90}{HallusionBench} & \rotatebox{90}{LLaVA-bench} & \rotatebox{90}{MMStar} & \rotatebox{90}{\textbf{Average}} \\
         \hline
         \hline  
    \modelname{} 7B  & 62.7 & 66.1 & 55.8 & 54.5 & \textbf{89.4} & \textbf{35.2} & \textbf{1543.5} & 316.4 & 70.3 & \textbf{46.8} & 65.0 & \textbf{36.8} & \textbf{58.3} \\
    Remove Caption Feature $\mathbf{V}_{t_1}'$   & 62.2 & 64.9 & 56.1 & 53.5 & 89.3 & 31.8 & 1477.8 & \textbf{354.3} & 69.0 & 44.9 & \textbf{65.2} & 36.0 & 57.6 \\
    Remove OCR Feature $\mathbf{V}_{t_2}'$   & 62.0 & 65.6 & 55.4 & 56.0 & 88.8 & 30.2 & 1506.3 & 345.4 & 67.6 & 45.4 & 62.6 & 35.2 & 57.3 \\
    Remove Grounding Feature $\mathbf{V}_{t_3}'$   & \textbf{63.0} & \textbf{66.6} & \textbf{56.8} & \textbf{56.5} & 88.8 & 32.9 & 1494.8 & 338.9 & \textbf{70.8} & 44.7 & 65.1 & 36.2 & 58.2 \\
        \hline
        \hline
        \end{tabular}
    \caption{Ablation study was conducted by removing one high level image feature at a time, demonstrating that all high-level features are essential for maintaining optimal performance.}
    \label{tab:feature_remove_ablation}
\end{table*}

\paragraph{Study on Breadth Features Impacts.}

 In Table \ref{tab:feature_remove_ablation}, we analyze the impact of each feature from different breadths by individually removing one feature at a time from $[\mathbf{V}_{t_1}', \mathbf{V}_{t_2}', \mathbf{V}_{t_3}']$. For instance, to assess the effect of the caption feature, we retain only the OCR and grounding features. The results in Table \ref{tab:feature_remove_ablation} show that combining all three features yields the best average benchmark performance.  
 Additionally, we plot the alignment loss when each feature is removed individually, as shown in Figure \ref{fig:feature_remove_loss}. This further demonstrates incorporating all three features from different breadths is essential for effectively extracting visual information.

\section{Related Work}
\label{sec:related}
LLMs have significantly advanced the development of MLLMs, including models like LLaVA \cite{liu2024visual}, MiniGPT-4 \cite{zhu2023minigpt}, Qwen-VL \cite{bai2023qwen}, and Vila \cite{lin2024vila}. Most of these models integrate a language-supervised vision encoder, such as CLIP or SigLIP, with a language model backbone. Beyond these, there is a wider range of visual models available, including self-supervised models \cite{oquab2023dinov2}, segmentation models \cite{kirillov2023segment}, and diffusion models \cite{rombach2022high}. Departing from conventional vision encoder designs, our work introduces an innovative approach by using the generative vision foundation model Florence-2 as the vision encoder.

While other studies, such as Cambrian \cite{tong2024cambrian}, Brave \cite{kar2024brave} and MouSi \cite{fan2024mousi} have explored the advantages of combining multiple visual signals, our approach avoids the added complexity and cost of using multiple vision encoders. Instead, we use a single vision model to generate multiple visual features,  which each one emphasizing different perceptual information in the input image. This approach allows us to achieve superior performance with a single vision encoder, surpassing models that rely on multiple vision encoders, such as Cambrian \cite{tong2024cambrian}.

High-resolution adaptation is commonly applied to increase the input resolution for MLLMs \cite{liu2024llavanext}. Besides, models like LLaVA-NeXT \cite{liu2024llavanext} and InternVL \cite{chen2024far} achieve this by using tiling or adaptive tiling, dividing high-resolution inputs into smaller patches for separate processing. Although our method does not incorporate these techniques, both approaches are compatible and could be combined with our method.


\section{Conclusion}
\label{sec:conclusion}
In conclusion, \modelname{} uses Florence-2 as a versatile vision encoder, which provides diverse, task-specific visual representations across multiple computer vision tasks like captioning, OCR, and Grounding. By leveraging Depth-Breadth Fusion (DBFusion), we incorporate a range of visual features from different layers ("Depth") and prompts ("Breadth") to create enriched representations that meet varied perceptual demands of downstream tasks. Our fusion strategy, based on channel concatenation, effectively combines these diverse features, which are then projected as input to the language model.

Through training on a novel data recipe that includes detailed captions for pretraining and diverse instruction tuning data, \modelname{} demonstrates superior alignment between the vision encoder and the LLM, outperforming other models across 25 benchmarks covering vision-centric, knowledge-based, and OCR \& Chart tasks. Our analysis underscores the effectiveness of Florence-2's generative capabilities in enhancing MLLM alignment and versatility for a wide range of applications. 

For future work, several avenues could further enhance the capabilities and efficiency of \modelname{}. One direction involves improving the DBFusion strategy by exploring more sophisticated fusion techniques that could dynamically adapt the Depth-Breadth balance based on specific downstream task requirements. Additionally, while Florence-2 provides diverse visual representations, future research could explore adaptive vision encoders that select features on-the-fly, optimizing computational efficiency without compromising performance.



\clearpage
\setcounter{page}{1}
\maketitlesupplementary

\section{Training Details}
 We selected two language backbones: Phi-3.5-mini-Instruct \footnote{https://huggingface.co/microsoft/Phi-3.5-mini-instruct} and LLama-3.1-8B-Instruct \footnote{https://huggingface.co/meta-llama/Meta-Llama-3.1-8B-Instruct}. For the main results, using the 16.9M image caption dataset and 10M instruction datasets, we trained all models on 8 nodes with 64 Nvidia H100 GPUs. The training process consists of two stages: pretraining and instruction tuning. During the pretraining stage, unlike LLaVA 1.5 which only tunes the projection layer, we fine-tune the entire model, including the vision backbone Florence-2, projection layer, and language model. We found that tuning the entire model yields better performance than freezing the vision and language models. In the fine-tuning stage, we tune only the projection layer and language models. For LLama-3.1-8B-Instrcut, the global batch size for pretraning stage is 256, with a cosine decay learning rate with maximun value 2e-5. In the fine-tuning stage, we maintain a global batch size of 256 and a learning rate of 1e-5. For Phi-3.5-mini-Instruct, the global batch size for pretraning stage is 4096, with a cosine decay learning rate with maximun value 1e-4. In the fine-tuning stage, the global batch size is 2048 and learning rate is 9e-5.

\section{Discussion}

\paragraph{OCR feature is essential for text based image understanding.}

In Table \ref{tab:ocr_ablation}, we examine the role of OCR in understanding images containing text. To evaluate the effect of the OCR feature, we retain only the caption and grounding features. The results in Table \ref{tab:ocr_ablation} indicate that, apart from TextVQA benchmark, the OCR feature is beneficial for extracting textual information from images in the other benchmarks.

\paragraph{Knowledge based benchmark reply more on the capability of language model.} In Table \ref{tab:knowledge_ablation} we removing the caption and grounding features does not result in a significant difference, suggesting that the knowledge-based benchmark scarcely relies on various visual information. Additionally, Table \ref{tab:combined_benchmarks} shows that the performance of the knowledge-based benchmark improves with the use of stronger language models.

\begin{table*}[ht]
    \centering
    \begin{subfigure}[b]{0.45\textwidth}
        \centering
        \begin{tabular}{c|ccccc}
         & \rotatebox{90}{OCRBench} & \rotatebox{90}{ChartQA} & \rotatebox{90}{DocVQA} & \rotatebox{90}{InfoVQA} & \textbf{\rotatebox{90}{Average}}\\
         \hline
         \hline	
        \modelname{} 7B &  \textbf{41.4} & \textbf{24.3} & \textbf{44.5} & \textbf{29.4} & \textbf{34.9}\\   
        OCR         &  40.9 & 22.9 & 44.4 & 29.0 & 34.2 \\
         \hline
         \hline
        \end{tabular}%
        \caption{Ablation study on OCR features on OCR \& Chart benchmark.}
        \label{tab:ocr_ablation}
    \end{subfigure}
    \hfill
    \begin{subfigure}[b]{0.45\textwidth}
        \centering
        \begin{tabular}{c|ccccc}
        & \rotatebox{90}{AI2D} & \rotatebox{90}{MathVista} & \rotatebox{90}{MMMU} & \rotatebox{90}{SciQA-IMG} & \rotatebox{90}{\textbf{Average}} \\
         \hline
         \hline	
        \modelname{} 7B  & \textbf{57.2} & \textbf{28.0} & 35.6 & \textbf{66.5} & 46.8 \\
        Caption &  56.8 & 27.5 & \textbf{36.9} & 65.5 & 46.7 \\
        OCR  & 55.7 & 27.0 & 35.8 & 65.6 & 46.0 \\
        Grounding &  56.7 & 27.9 & \textbf{36.9} & 66.4 & \textbf{47.0} \\
         \hline
         \hline
        \end{tabular}%
        \caption{Ablation Studies on Knowledge based benchmarks.}
        \label{tab:knowledge_ablation}
    \end{subfigure}
    \caption{Ablation studies on different features for various benchmarks.}
    \label{fig:ocr_knowledge_ablation}
\end{table*}

{
    \small
    \bibliographystyle{ieeenat_fullname}
    \bibliography{main}
}


\end{document}